# FEC: Three Finetuning-free Methods to Enhance Consistency for Real Image Editing

Songyan Chen[1,2,a], Jiancheng Huang[1,2,a,*]
[1]*Shenzhen Institute of Advanced Technology, Chinese Academy of Sciences*
[2]*University of Chinese Academy of Sciences*
{sy.chen, jc.huang}@siat.ac.cn

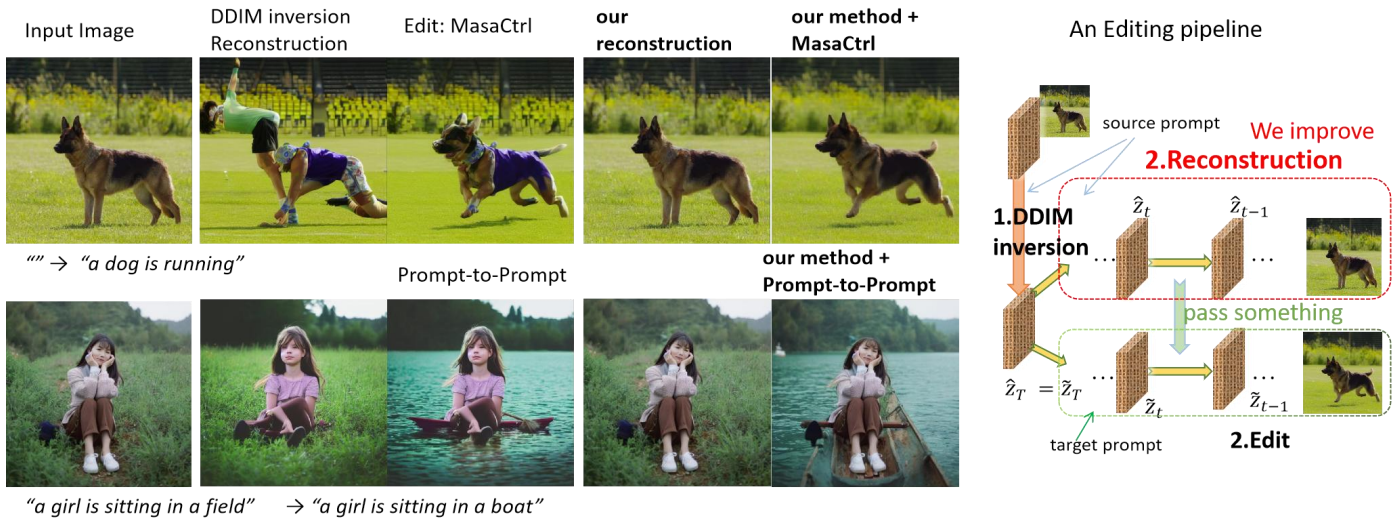

Fig. 1: Our methods help **maintain consistency of unedited parts** in real image editing. Real image editing under text-to-image diffusion model typically consists of stages: DDIM inversion, fine-tuning (optional), reconstruction and editing (see right figure). Distorted reconstruction(column 2) may lead to inconsistent editing outcomes(column 3). Our methods, without fine-tuning, can reconstruct successfully(column 4), thus passing correct information to the editing route. Therefore, the dog/girl in the edited outcome (column 5) is identical to the original one(column 1).

*Abstract*—Text-conditional image editing is a very useful task that has recently emerged with immeasurable potential. Most current real image editing methods first need to complete the reconstruction of the image, and then editing is carried out by various methods based on the reconstruction. Most methods use DDIM Inversion for reconstruction, however, DDIM Inversion often fails to guarantee reconstruction performance, i.e., it fails to produce results that preserve the original image content. To address the problem of reconstruction failure, we propose FEC, which consists of three sampling methods, each designed for different editing types and settings. Our three methods of FEC achieve two important goals in image editing task: 1) ensuring successful reconstruction, i.e., sampling to get a generated result that preserves the texture and features of the original real image. 2) these sampling methods can be paired with many editing methods and greatly improve the performance of these editing methods to accomplish various editing tasks. In addition, none of our sampling methods require fine-tuning of the diffusion model or time-consuming training on large-scale datasets. Hence the cost of time as well as the use of computer memory and computation can be significantly reduced.

## I. Introduction

With the hot trend of generative large-scale models, many generative model-based approaches are attracting great attention from researchers. In particular, text-conditioned image generation is very hot in both the research and commercial worlds, such as the familiar AI painting and AI image creation, both of which are often experienced by people in their everyday lives [1]–[5]. However, text-conditioned AI painting or generating images is not enough to satisfy everyone's needs, and many users want to be able to edit the real images they provide by using text as a condition. Thus, diffusion model-based image editing methods have exploded in recent months, leading to rapid and obvious breakthroughs in image editing tasks. Its potential for successful application in comic book production, video editing, advertising material production is immeasurable. However, real image editing based on diffusion model still has obvious problems and great challenges.

In order to understand the existing problems, the process and the fundamental principle of text-conditioned generative model of images based on diffusion model is first briefly described. At the beginning, the user provides a text prompt $P_s$ (e.g., a photo of a dog), and a Gaussian noise named $x_T$ is randomly sampled, then the generative model will start its sampling process with this $x_T$ as the starting point. The sampling process has multiple steps (generally 50 steps in commercial use). The diffusion model generates new content in the noised latent at each sampling step based on the given text prompt $P_s$. Finally, a image of a dog can be generated.

For real image editing, the first thing that needs to be guaranteed is that the real image can be reconstructed during editing; if it can't be reconstructed at all, then there is no point in editing, and reconstruction performance is the upper limit of editing performance. For the reconstruction of the real image, existing methods mostly utilize DDIM inversion [6]. DDIM Inversion can obtain a noised latent $x_T$ from a real image $I_s$ by inversion of the sampling process and use this $x_T$ as a starting point for the sampling process. Then, a reconstructed image $\hat{I}_s$ can theoretically be obtained by the above sampling process. With DDIM Inversion, many image editing methods first invert the real image into $x_T$, and then generate different content (i.e. image editing) by modifying the trajectory of $x_t$ during sampling.

However, a lot of work has indicated that the problem of reconstruction failure is very common, basically when the guidance scale is greater than 1 the reconstruction fails, and some images can't be reconstructed even when the guidance scale is set to 1 or below. The experiments analysing the sensitivity to this guidance scale are detailed in Sec. IV-A3. Failure of reconstruction is a very serious problem, because reconstruction failure means that it is difficult to maintain the consistency between the editing result and the original image. This phenomenon also occurs in other works [2], [7]–[9].

In order to solve the problem of reconstruction failure, some methods choose to finetune certain parameters, such as Null-Text Inversion [10], which makes the reconstructed trajectory and the inversion trajectory aligned as much as possible by training the unconditional embedding in the finetune stage, however, the method needs to consume time in this finetuning. Therefore Negative-prompt Inversion [11] and Proximal Negative-Prompt Inversion [12] are successively proposed, they prove that the optimisation approach of Null-Text Inversion [10] is unnecessary and proposed their optimisation-free approach to help reconstruction, however Negative-Prompt Inversion actually just degenerates to setting the guidance scale to 1, and Proximal Negative-Prompt Inversion requires controlling the range space of the optimisation result through artificially specified hyperparameters. These optimisation-free methods can achieve good reconstruction performance, however they have drawbacks, such as they need to occupy the position of the negative prompt, resulting in some editing methods [7] under which the user cannot use the negative prompt.

In order to solve the reconstruction problem better and more conveniently under the training-free setting, we first analyse the reason for the reconstruction failure from the perspective of mathematical theory, which is the difference between inversion and sampling. Then, we propose FEC, which consists of three sampling methods that guarantee excellent reconstruction performance under different settings. Three sampling methods of FEC are FEC-ref, FEC-noise and FEC-kv-reuse, and they can handle different types of editing tasks. In this paper, we basically divide all editing tasks into two classes, rigid and non-rigid editing.

Rigid editing usually refers to the editing result has the same overall structure with the source image, that is to say, the SSIM of their edge maps is high. This type of editing includes object replacement, object adding, scene replacement, style transfer and so on. Non-rigid editing involves modifying the structure of an object, such as editing posture (action) of animals and people. Next we detail which editing tasks the three sampling methods can be used for.

1) FEC-ref means to save the intermediate noisy latent variable as a reference path during inversion, and then use this reference path to directly guide our editing path during sampling to ensure the consistency of the reconstruction effect and editing. FEC-ref can be used on both rigid and non-rigid editing.

2) FEC-noise means that we calculate the desired noise at each step during sampling according to current noisy latent and the target latent. Since the nature of the predicted noise during sampling is the change trajectory of the latent image, we can ensure that the changes in the non-edited regions of the editing path are in the direction of the same content of the source image by calculating the desired noise in the non-edited regions during sampling. FEC-noise can be used on rigid editing.

3) FEC-kv-reuse firstly saves the Key and Value embedding in the self-attention layer at each step of UNet during Inversion stage. After obtaining these KV embedding, if we need to reconstruct the image, we can rely on these embedding to help correct the error during sampling, so that the trajectory of sampling is symmetric with the trajectory of inversion, thus achieving good reconstruction performance. When we need to edit the image, we can modify the prompt as the target prompt for editing, the target prompt can introduce new semantic information for UNet, and combined with the saved KV embedding mentioned above, we can achieve the image editing with consistent content. FEC-kv-reuse can be used on non-rigid editing.

In summary, our contributions are as follows:

- In order to solve the problem of reconstruction failure of Stable Diffusion on real images, we propose FEC including 3 sampling methods, which can achieve very good reconstruction performance without training, and can also avoids the parallelism bug that often occurs when the batch-size is larger than 1 in the diffusion model.
- Our 3 sampling methods in FEC can also help the image editing task very well, in the same mode as the reconstruction, it only requires the user to change the edit prompt to finish the editing, and no training or finetuning is needed.
- Qualitative and quantitative experiments on two benchmarks show that our method can achieve very good reconstruction performance and satisfactory editing results with the lowest memory usage and minimum time.

## II. RELATED WORK

### A. Text-to-Image Generation and Editing

With the hotness of diffusion models, text-conditioned image generation models have experienced an unprecedented explosion [4], [5], [13]–[16]. Diffusion model is a generative model designed based on the laws of physics, and its general pattern is to add Gaussian noise associated with a time step on top of a data distribution (e.g., an image), and then use a noise estimation network in the training stage to predict those added noises. At the sampling stage after training (which can be interpreted as a testing stage), the trained noise estimation network can start with a pure Gaussian noise, gradually remove the noise and generate new content until a complete new image is finally generated.

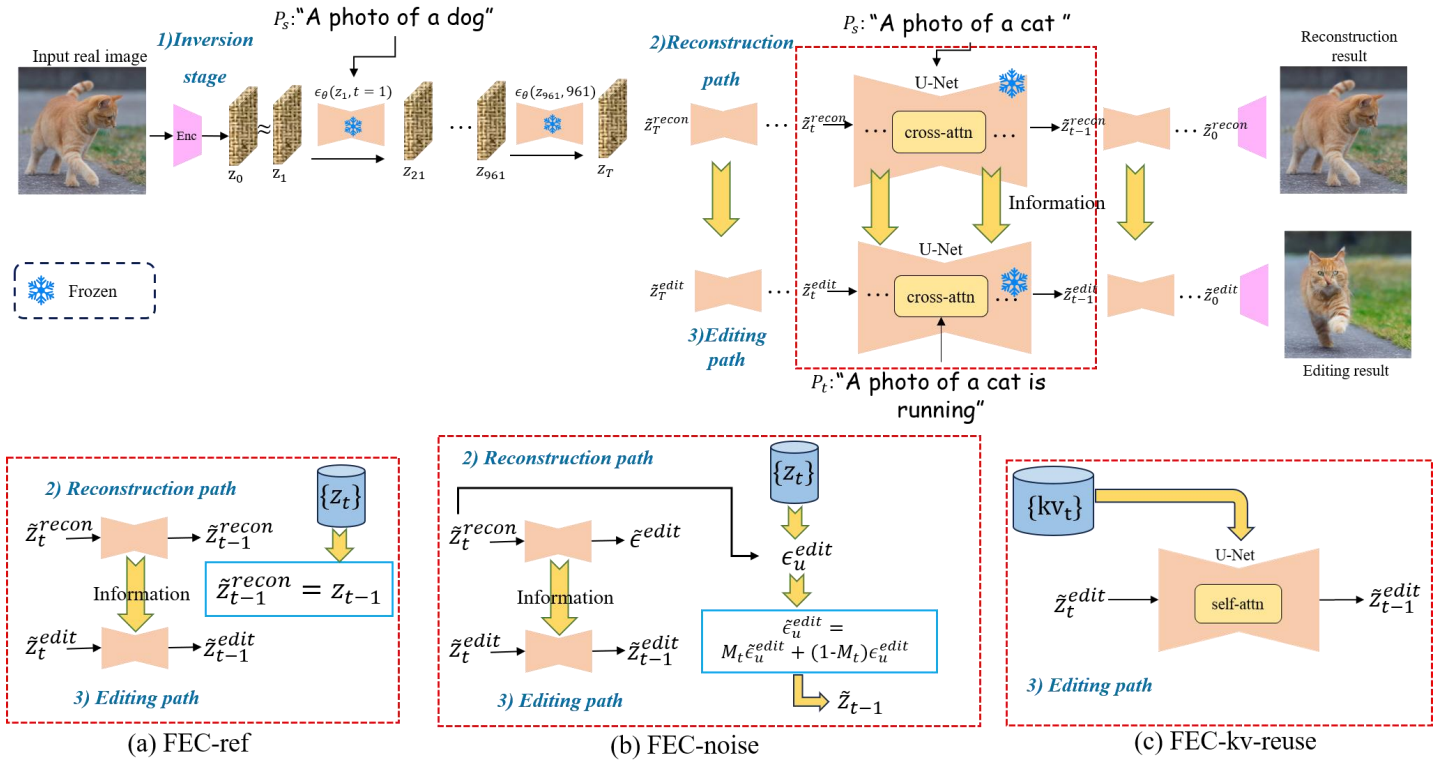

Fig. 2: Pipeline of the proposed 3 sampling methods of FEC. Each method has inversion stage and sampling stage. Inversion stage is for getting noisy latent and some valuable information. Sampling stage is using the noisy latent and some valuable information for consistency editing results.

Text-conditioned image generation models based on diffusion models generally match textual prompt to image content by using pretrained multimodal visual language models, such as CLIP [17]. The introduction of textual prompt has provided diffusion models with superior and diverse generative capabilities, and has led to cutting-edge results in terms of image quality and diversity, including DALL·E 2 [4], LDM [5], VQ-Diffusion [18], InstructPix2Pix [19] and GLIDE [2], among others, which generate high-quality images consistent with a given textual prompt. On the basis of image generation, some text-based image editing methods such as P2P [7], InstructPix2Pix [19], PnP [8], and MasaCtrl [20] are beginning to demonstrate their editing capabilities, and another class of subject-driven image generation methods such as Textual Inversion [21], Dreambooth [22], Custom Diffusion [23], ELITE [24], and FasterComposer [25] have also been proposed. Many of them rely on DDIM Inversion [6] for real image editing, however, as in Sec. I, the reconstruction failure of DDIM Inversion often leads to editing failure as well.

### B. Stable Diffusion Model

FEC is designed for Stable Diffusion (SD) [5]. In Stable Diffusion, in order to solve the problem of high-resolution image generation, VAE's autoencoder is used to encode the natural image $x_0$ into the $z_0$ of the hidden space, and decode the $z_0$ back to the $x_0$, which has the advantage of reducing the amount of computation and the runtime significantly. In addition, the U-Net [26] is used as a noise estimation network, generally denoted by $\epsilon_\vartheta$, which is responsible for predicting the noise contained in $z_t$. The U-Net $\epsilon_\vartheta$ generally consists of a basic Resblock and a Transformer block, where the Transformer block contains self-attention and cross-attention layers. According to some pre-experiments, it was found that cross-attention is mainly responsible for introducing the information from text embedding into the U-Net network, thus providing the overall semantics (and also part of the texture) and structure of $z_t$ [7], [8], [27]. Self-attention, on the other hand, is responsible for interacting the image token, and thus it is mainly responsible for the specific texture details. Specifically query $Q$ is generally derived from image features, while key $K$ and value $V$ are derived from text features or image features themselves. Based on the previous experiments, we know that image generation is affected by several factors: first, the attention map of cross-attention plays a crucial role in the overall structure, such as the location of objects. Second, the texture and details of the generated image are attributed to the operation of $K$ and $V$ in the cross-attention and self-attention layers.

### C. Preliminaries

Text-guided diffusion models aim to map a random noise vector $z_t$ and textual condition $P$ to an output image $z_0$, which corresponds to the given conditioning prompt. In order to perform sequential denoising, the network $\epsilon_\vartheta$ is trained to predict artificial noise, following the objective:

$$\min_{\vartheta} E_{z_0, \epsilon \sim N(0,I), t \sim \text{Uniform}(1,T)} \| \epsilon - \epsilon_\vartheta(z_t, t, C) \|_2 \quad (1)$$

Note that $C = \psi(P)$ is the embedding of the text condition and $z_t$ is a noised sample, where noise is added to the sampled data $z_0$ according to timestamp $t$. At inference, given a noise vector

$z_T$, The noise is gradually removed by sequentially predicting it using our trained network for $T$ steps.

Since we aim to accurately reconstruct a given real image, we employ the deterministic DDIM sampling [6]:

$$z_{t-1} = \sqrt{\alpha_{t-1}} \frac{z_t - \sqrt{1-\alpha_t} \cdot \epsilon_\vartheta(z_t, t)}{\sqrt{\alpha_t}} + \sqrt{1-\alpha_{t-1}} \cdot \epsilon_\vartheta(z_t, t),$$

For the definition of $\alpha_t$ and additional details, please refer to [6]. Diffusion models often operate in the image pixel space where $z_0$ is a sample of a real image. In our case, we use the popular and publicly available Stable Diffusion model [28] where the diffusion forward process is applied on a latent image encoding $z_0 = E(x_0)$ and an image decoder is employed at the end of the diffusion backward process $x_0 = D(z_0)$.

*a) Classifier-free guidance.:* One of the key challenges in text-guided generation is the amplification of the effect induced by the conditioned text. To this end, Ho et al. [29] have presented the classifier-free guidance technique, where the prediction is also performed unconditionally, which is then extrapolated with the conditioned prediction. More formally, let $\varnothing = \psi("")$ be the *embedding* of a null text and let $w$ be the guidance scale parameter, then the classifier-free guidance prediction is defined by:

$$\tilde{\epsilon}_\vartheta(z_t, t, C, \varnothing) = w \cdot \epsilon_\vartheta(z_t, t, C) + (1-w) \cdot \epsilon_\vartheta(z_t, t, \varnothing).$$

E.g., $w = 7.5$ is the default parameter for Stable Diffusion.

*b) DDIM inversion.:* A simple inversion technique was suggested for the DDIM sampling [6], [30], based on the assumption that the ODE process can be reversed in the limit of small steps:

$$z_t \approx \sqrt{\alpha_t} \frac{z_{t-1} - \sqrt{1-\alpha_{t-1}} \cdot \epsilon_\vartheta(z_{t-1}, t)}{\sqrt{\alpha_{t-1}}} + \sqrt{1-\alpha_t} \cdot \epsilon_\vartheta(z_{t-1}, t).$$

In other words, the diffusion process is performed in the reverse direction, that is $z_0 \rightarrow z_T$ instead of $z_T \rightarrow z_0$, where $z_0$ is set to be the encoding of the given real image.

## III. METHOD

We first introduce our task in Sec. III-A. Then we analyse the reasons for the failure of reconstruction through DDIM inversion and DDIM sampling in Sec. III-B, and then logically propose the three methods of FEC with their own inversion stage and sampling stage, respectively, to theoretically explain why our reconstruction and editing can be successful.

### A. Task Setting

Given a real rather than a synthetic source image $I_s$ and a corresponding text prompt $P_s$ (which can be semantic text or empty), the goal of our task is to synthesize a new ideal image $I_t$ with a pre-trained Stable Diffusion Model that matches the target editing text prompt $P_t$. This editing result image $I_t$ should meet the following requirements: 1) $I_t$ semantically matches the text prompt of $P_t$, e.g., it can satisfy that the corresponding object is performing the corresponding action. 2) The object inside $I_t$ should be consistent with $I_s$ in terms of content. For example, given a real image (corresponding to $I_s$) with a cat standing still, we edit the text prompt $P_t$ to "a running cat", and then generate a new image with the cat running, while the other contents of the image remain basically unchanged.

This task is very difficult, especially when used on real images. Even using $\hat{z}_T$ as the starting point of the reverse process obtained from DDIM Inversion of the original real image $I_s$, the reconstruction image $\hat{I}_s$ is often very different from the original real image $I_s$ [7].

### B. Overall Motivation

*1) Failure of DDIM Inversion Reconstruction:* Linear cumulative error is a significant cause of reconstruction failure in DDIM Inversion. Starting with the fundamentals, we first review the training process of Stable Diffusion [5]. Each time a time step $t \in [1, T]$ ($T$ equals 1000 in most cases) is first sampled and then random Gaussian noise $\epsilon \in N(0, I)$ is added to the level $t$ using the forward process equation:

$$\mathbf{z}_t = \sqrt{\alpha_t} \mathbf{z}_0 + \sqrt{1-\alpha_t} \epsilon. \quad (2)$$

Note that $t$ represents the level of noise instead of number of times the noise is added. Then, $\mathbf{z}_t$ and $t$ are passed through the noise estimation network $\epsilon_\vartheta(\mathbf{z}_t, t)$ to predict the added noise $\epsilon$. Therefore, MSE loss function is applied for the training of the network $\epsilon_\vartheta(\mathbf{z}_t, t)$:

$$L_{\text{simple}} = |\epsilon - \epsilon_\vartheta(\mathbf{z}_t, t)|. \quad (3)$$

Having introduced the training, we now move on to the analyses of DDIM Inversion. DDIM Inversion derives from DDIM sampling. With $\tilde{z}_t$ denoting the noisy latent, DDIM sampling is as:

$$\tilde{z}_{t-1} = \sqrt{\alpha_{t-1}} \frac{\tilde{z}_t - \sqrt{1-\alpha_t} \cdot \epsilon_\vartheta(\tilde{z}_t, t)}{\sqrt{\alpha_t}} + \sqrt{1-\alpha_{t-1}} \cdot \epsilon_\vartheta(\tilde{z}_t, t), \quad (4)$$

$$\epsilon_\vartheta(\tilde{z}_t, t) = \omega \epsilon_\vartheta(\tilde{z}_t, t, C) + (1-\omega) \epsilon_\vartheta(\tilde{z}_t, t, \varnothing), \quad (5)$$

where $\omega$ denotes the guidance scale of classifier-free guidance (CFG), and $C$, $\varnothing$ represent text embedding of prompt and negative-prompt, respectively. Negative-prompt is also known as unconditional embedding. Assuming that there exists an approximation $\epsilon_\vartheta(\tilde{z}_t, t) \approx \epsilon_\vartheta(\tilde{z}_{t-1}, t)$ when the interval between $t$ and $t-1$ is sufficiently small, then can drive DDIM inversion formula by the inversion of Eq. 4 as:

$$z_t \approx \sqrt{\alpha_t} \frac{z_{t-1} - \sqrt{1-\alpha_{t-1}} \cdot \epsilon_\vartheta(z_{t-1}, t)}{\sqrt{\alpha_{t-1}}} + \sqrt{1-\alpha_t} \epsilon_\vartheta(z_{t-1}, t). \quad (6)$$

Since DDIM Inversion exists $\epsilon_\vartheta(z_t, t) \neq \epsilon_\vartheta(z_{t-1}, t)$, there is an assumption which is subject to error. Since the noise level of $\mathbf{z}_{t-1}$ is different from that of $t$, the noise estimation network $\epsilon_\vartheta$ faces a situation that has never been seen during training. In particular, most practical scenarios, DDIM uses skip sampling, e.g. "$t$" = 981 and "$t-1$" = 961 (the interval between skips is 20 time steps), which further amplifying the above problem.

$$||\tilde{z}_{t-1} - z_{t-1}|| \propto ||\epsilon_\vartheta(\tilde{z}_t, t) - \epsilon_\vartheta(z_{t-1}, t)|| \quad (7)$$

$$||\tilde{z}_{t-2} - z_{t-2}|| > ||\tilde{z}_{t-1} - z_{t-1}|| \quad (8)$$

Since $\tilde{z}_{t-1}$ at reconstruction is already in error with $z_{t-1}$ (Eq. 7), and the next prediction of the noise $\epsilon_\vartheta(\tilde{z}_t, t)$ requires the use of this erroneous $\tilde{z}_{t-1}$, this leads to a larger error (linear cumulative error) for $\tilde{z}_{t-2}$ and $z_{t-2}$ than $\tilde{z}_{t-1}$ and $z_{t-1}$. During the sampling of the reconstruction, tens to hundreds of steps of the denoising process are required, and thus the error of the latent gradually accumulates, leading to reconstruction failures, especially when an inappropriate prompt is used, or the guidance scale exceeds 1. This means that the editing process continuously introduces wrong information, resulting in editing failure. The experimental reasons why our FEC can address this failure are detailed in Sec. IV-A2.

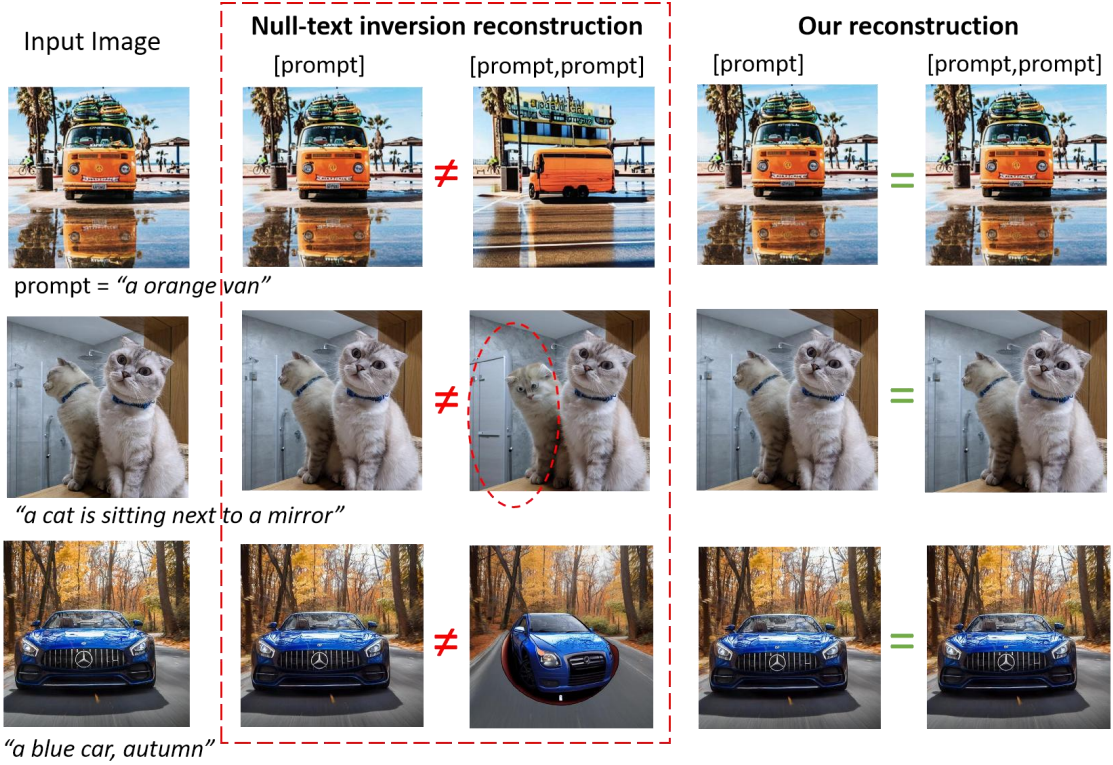

Fig. 3: Problem of misaligned batch size. y1 = f (x), [y2, y2] = f ([x, x]), ||y1 − y2 || > 0 in PyTorch. In Null-text inversion [31], editing(2 prompts) uses twice the batch size of Text Encoder and UNet compared to tuning stage(1 prompt). Editing may sometimes reconstruct fail in details, as shown in the red box. Our methods can avoid this problem.

*2) The Problem of Parallel Input:* An intuitively naive way to solve the above problem of reconstruction failure is to use parallel inputs at sampling, i.e., batch-size of 2. However, this introduces a new problem, which we analyse in this paragraph.

Stable Diffusion has a large difference in output on **serial vs. parallel** input. U-Net, text encoder, and other components of Stable Diffusion suffer from this problem:

$$y_1 \leftarrow SD(x), \quad [y_2, y_2] \leftarrow SD([x, x]), \quad (9)$$

$$||y_1 - y_2|| > 10^{-5}, \quad (10)$$

where $SD$ denotes the Stable Diffusion Model and $x$ is any input. After tens or hundreds of computationally cumulative steps, the final results of serial and parallel may reach a difference that is noticeable to the naked eye, as shown in Fig. 3. We found this to be due to a bug in the specific implementation of Pytorch's Linear layer, Fig. 3 demonstrates the same phenomenon. Despite the presence of this noise, state-of-art editing methods (e.g. Null-Text + prompt2prompt [10], Masactrl) are all in parallel to perform image reconstruction and editing. This parallel bug can lead to unsatisfactory editing results. And if the reconstruction and editing are serial, while it solves the problem, it increases the reasoning time at generation time (by about 8s). We show the detail experimental results about batchsize in Sec. IV-A5.

*3) Unnecessary GPU Memory Usage:* Redundant steps in some editing methods [20] waste time and memory. Masactrl editing requires a self-attention Key and Value embeddings of $\epsilon_\vartheta(z_t, t)$ at the reconstruction path. If we could save this information at inversion stage, we could dispense with the reconstruction path, saving time and GPU memory, and not having problems with the above parallel input.

---

**Algorithm 1** The 2 Stages of FEC-ref

**Require:** the latent of the original real image $z_0$ and the source prompt embeddings $c_s$. The initialization weights $\alpha = \{\alpha_t, t = 1, ..., T\}$.

---

**1) Inversion stage:** Set guidance scale $w = 7.5$ of stable diffusion model, in this stage we utilize DDIM inversion to obtain the intermediate latents$\{z_t, t = 1, ..., T\}$.

**2) Sampling stage:** Set guidance scale $w = 7.5$ and the beginning of reverse process $\tilde{z}_T = z_T$. Given a editing prompt $P_t$, then get its embedding $c_t$ and an unconditional embedding $c_u$;

**for** $t = T, T − 1, . . . , 1$ **do**
  $\epsilon_c = \epsilon_\vartheta(\tilde{z}_t, t, c_c); \epsilon_u = \epsilon_\vartheta(\tilde{z}_t, t, c_u); \tilde{\epsilon}_t = \tilde{\epsilon}_u + w(\tilde{\epsilon}_c − \tilde{\epsilon}_u);$
  If reconstruct: $\tilde{z}_{t−1} = z_{t−1};$
  Else edit: $\tilde{z}_{t−1}$ = PrevStep$(\tilde{z}_t, \epsilon_t)$ as Eq. 4;
**end**
$I_t$ = Decode($\tilde{z}_0$);
**Return** Editing result $I_t$

---

### C. FEC-ref

*1) Implementation Motivation:*

In order to alleviate linear cumulative error introduced in Sec. III-B, we first choose to directly utilize the latents during inversion and propose FEC-ref. Specifically, as illustrated in Fig. 2 (a), FEC-ref means using the trace of inversion (the noised latents $z_t$) as reference latents to correct the linear cumulative error producing during each step of sampling.

*2) Implementation Detail:*

We first perform DDIM inversion to get a series of latent noises $z_t$. Now, we define the saved reference latents as $Z_{ref} = [z_1, ..., z_T]$. With these reference latents as $Z_{ref}$, we define FEC-ref Sampling as:

$$\tilde{z}_{t-1} = z_{t-1}, \quad (11)$$

where $\tilde{z}_{t-1}$ and $z_{t-1}$ denote the noise latent of the reconstruction path during sampling and the noise latent during inversion, respectively. Fig. 2 (a) shows that this process. By Eq. 11, we use these reference latents $z_{t-1}$ to correct the error in $\tilde{z}_{t-1}$ at each step during sampling and reduce the above linear cumulative error, which can make sure a perfect reconstruction.

After guaranteeing the quality of the reconstruction, editing task can be completed by combining with common editing methods such as P2P and MasaCtrl. For example, when combining with P2P, we use FEC-ref on the reconstructed path of P2P, so that we can correct the linear cumulative error of P2P's reconstructed path. Then, P2P can deliver the accurate information in the reconstructed path such as the cross-attention map to the editing path. The meaning of FEC-ref is to help the reconstruction path to achieve better results, so that the reconstruction path can deliver more accurate information to the editing path.

### D. FEC-noise

*1) Implementation Motivation:*

For rigid editing such as object replacement, we analyse the denoising path in terms of the predicted noise, and propose FEC-noise. Because we know that the nature of the predicted noise at each step during the sampling process implies a change in the direction of the sampling path, it is important to analyse the desired change (the predicted noise) at each step during the sampling, which can have huge impact on the direction for perfect reconstruction.

*2) Implementation Detail:*

Specifically, as illustrated in Fig. 2 (b), we also perform DDIM inversion to save a series of latent noises $z_t$.

As for how to compute the desired noises $\epsilon_t$ to correct the sampling path, we consider that the unedited part of the target image should be same as the corresponding part in the source image. Because predicted noise means the direction of the sampling path, we can easily use the saved next target latent $z_{t-1}$ to compute the desired noise $\epsilon_t$ as:

$$\epsilon_t = \frac{\sqrt{\alpha_t}\tilde{z} - \sqrt{\alpha_{t-1}}z}{\sqrt{\alpha_t(1-\alpha^{t-1})} - \sqrt{\alpha^{t-1}(1-\alpha_t)}}, \quad (12)$$

$$\epsilon_u = (\epsilon_t - \omega\tilde{\epsilon}_c)/(1-\omega), \quad (13)$$

where the desired noise $\epsilon_t$ represents the direction from $\tilde{z}_t$ to $z_{t-1}$. $\tilde{\epsilon}_c$ and $\epsilon_u$ denote the conditional and unconditional predicted noises. With the desired $\epsilon_u$, we can make the sampling latent become more and more like the source image for perfect reconstruction.

However, for editing task, we should only reconstruct the unedited part. Thus, we are faced with the problem of quantitatively defining the unedited part. Following [7], we can provide a blend word denoting the editing object such as "cat", and then extract the average cross-attention map of the token "cat" in each step as the segment mask of editing part $M \in \{0,1\}^{H \times W}$, which indicates the area to be edited.

With $M_t$ from each sampling step, we define FEC-noise Sampling as:

$$\tilde{\epsilon}_u = M_t\tilde{\epsilon}_u + (1 - M_t)\epsilon_u, \quad (14)$$

where $\tilde{\epsilon}_u$ and $\epsilon_u$ denote the predicted noise of the editing path during sampling and the computing desired noise, respectively. Since $M_t$ is the mask of editing part, $(1 - M_t)$ means the complementary part of the image. By using the above method, we can introduce desired noise for construction at each step to fill in the non-edited part, making the non-edited part change towards the content of the source image, so as to achieve consistent reconstruction or editing. Fig. 2 (b) shows that this process.

For better reconstruction effect, we use $M_t$ in $t = [T, 1]$, i.e., replace most of the non-edited part of the editing path during sampling with the desired noise.

### E. FEC-kv-reuse

*1) Implementation Motivation:*

As mentioned in Sec. I, the features in self-attention layers such as Key and Value have a huge impact on the content and texture of the generated object. So if we need a special design for non-rigid editing, Key and Value in the self-attention layer are the most important to be considered. Based on [20], we know that replacing the Key and Value in self-attention layer with the corresponding ones during inversion has a benefit on non-rigid editing. Thus, we propose FEC-kv-reuse.

*2) Implementation Detail:* Specifically, as illustrated in Fig. 2 (c), for obtaining the Key $K$ and Value $V$ in the self-attention layer of $\epsilon_\vartheta(z_t, t)$ during the trace of $z_t$, we first perform inversion to synthesize a series of latent noises $\{z_t\}$ including $z_T$.

Now, we define an inversion formula:

$$z_t = \frac{\sqrt{\alpha_t}z_{t-1} - \sqrt{1-\alpha_{t-1}} \cdot \bar{\epsilon}_\vartheta(z_{t-1}, t-1)}{\sqrt{\alpha_{t-1}}} + \sqrt{1-\alpha_t} \cdot \bar{\epsilon}_\vartheta(z_{t-1}, t-1). \quad (15)$$

where $\bar{\epsilon}_\vartheta(z_t, t)$ means that the $K$ (Key) and $V$ (Value) of the self-attention layer are preserved when passed through the $\epsilon_\vartheta$. We denote $KV_t = \{KV_{t,1}, ......, KV_{t,L}\}$, where $KV_{t,l} = (K_{t,l}, V_{t,l})$, and $l$ represent the $l$-th Transformer block. The reason for preserving the $KV_t$ is that it contains the content, texture, and identity of the noisy latents, and introducing $KV_t$ in the sampling stage keeps the corresponding generated image consistent with the original image $I_s$ in terms of content, texture and identity.

The input term $z_t$ in the noise estimation network $\bar{\epsilon}_\vartheta(z_t, t)$ in Eq. 15 is aligned with the time step $t$ instead of $\epsilon_\vartheta(z_{t-1}, t)$ in Eq. 6. The advantage of such alignment is that the $KV_t$ saved in our inversion can be used directly when sampling $\hat{\epsilon}_\vartheta(\tilde{z}_t, t)$, in this way making the inputs to the U-Net the same, for experimental analysis see the Sec. IV-A2.

For better reconstruct $I_s$, we define the FEC-kv-reuse self-attention block as follows. For instance, given the $l$-th self-attention layer during step $t$ of reverse process, we have:

$$\text{Attention}(\tilde{Q}_{t,l}, K_{t,l}, V_{t,l}) = \text{Softmax}(\frac{Q_{t,l}K_{t,l}'}{\sqrt{d}})V_{t,l}. \quad (16)$$

**Algorithm 2** The 2 Stages of FEC-noise
**Require:** the latent of the original real image $z_0$ and the source prompt embeddings $c_s$.

**1) Inversion stage:** Set guidance scale $w = 7.5$ of stable diffusion model, in this stage we utilize DDIM inversion to obtain the trace of noises $\{z_t, t = 1, ..., T\}$ and save the added noises $\{\epsilon_t, t = 1, ..., T\}$.

**2) Sampling stage:** Set guidance scale $w = 7.5$ and the beginning of reverse process $\tilde{z}_T = z_T$. Given a editing prompt $P_t$, then get its embedding $c_t$ and an unconditional embedding $c_u$;

for $t = T, T - 1, ..., 1$ do
    $\epsilon_c = \epsilon_\vartheta(\tilde{z}_t, t, c_c)$;
    Obtain $M_t$ from the cross-attention map (if available);
    **if** *reconstruct* **then**
        obtain $\epsilon_u^{recon}$ from $\epsilon_c, z_t, \tilde{z}_{t-1}$, as Eq. 12;
    **else if** *edit* **then**
        $\epsilon_u^{edit} = \epsilon_u^{recon}$,
        $\tilde{\epsilon}_u^{edit} = M_t \tilde{\epsilon}_u^{edit} + (1 - M_t)\epsilon_u^{edit}$;
    **end**
    $\tilde{\epsilon}_t = \tilde{\epsilon}_u + w(\tilde{\epsilon}_c - \tilde{\epsilon}_u)$;
    $\tilde{z}_{t-1}$ = PrevStep($\tilde{z}_t, \tilde{\epsilon}_t$) as Eq. 4;
end
$I_t$ = Decode($\tilde{z}_0$);
**Return** Editing result $I_t$

---

**Algorithm 3** The 2 Stages of FEC-kv-reuse
**Require:** the latent of the original real image $z_0$ and the source prompt embeddings $c_s$.

**1) Inversion stage:** Set guidance scale $w = 7.5$ of stable diffusion model, in this stage we utilize DDIM inversion to obtain the intermediate latents $\{z_t, t = 1, ..., T\}$ and self-attention Key and Value $kv_t$ at each step.

**2) Editing stage:** Set guidance scale $w = 7.5$ and the beginning of reverse process $\tilde{z}_T = z_T$. Given a editing prompt $P_t$, then get its embedding $c_t$ and an unconditional embedding $c_u$;

for $t = T, T - 1, ..., 1$ do
    Replace the $l_t$ layers of self-attention KV in model $\epsilon_\vartheta$ with the saved corresponding $kv_t$ :
    $\epsilon_c = \hat{\epsilon}_\vartheta(\tilde{z}_t, t, c_c; kv_t, l_t)$,    $\epsilon_u = \hat{\epsilon}_\vartheta(\tilde{z}_t, t, c_u; kv_t, l_t)$;
    $\tilde{\epsilon}_t = \tilde{\epsilon}_u + w(\tilde{\epsilon}_c - \tilde{\epsilon}_u)$;
    $\tilde{z}_{t-1}$ = PrevStep($\tilde{z}_t, \epsilon$) as Eq. 4;
end
$I_t$ = Decode($\tilde{z}_0$);
**Return** Editing result $I_t$

---

With this FEC-kv-reuse self-attention block, we define FEC-kv-reuse Sampling as:

$$\tilde{z}_{t-1} = \sqrt{\alpha_{t-1}} \frac{z_t - \sqrt{1-\alpha_t} \epsilon_\vartheta(z_t, t, kv_t, l_t)}{\sqrt{\alpha_t}} + \sqrt{1-\alpha_{t-1}} \cdot \hat{\epsilon}_\vartheta(\tilde{z}_t, t; kv_t, l_t), \quad (17)$$

where $\hat{\epsilon}_\vartheta(\tilde{z}_t, t; kv_t, l_t)$ denotes the noise estimation network $\epsilon_\vartheta$ using FEC-kv-reuse self-attention block within the range of the ordinal number $l_t = (l_{t,s}, l_{t,e})$ by the corresponding saved ones in $KV_t$. Fig. 2 (c) shows that this process.

To observe the reconstruction effect, let $l_t = (0, 16)$, i.e., replace all the Key and Value of the self-attention layer with the corresponding saved ones in $KV_t$. The reason why the reconstruction needs to use the saved $KV_t$ is that this brings in the original image's content, texture, and identity information well, and in the case of aligning the input $\tilde{z}_t$ and $t$, the use of $KV_t$ allows the reconstruction path to be similar enough to the inversion path. Besides, only using Value may also lead to poor performance as experiment in Sec. IV-C.

The FEC-kv-reuse sampling formula in Eq. 17 is consistent with DDIM sampling, and thus the forward process formula derived from it is consistent with the forward process formula in DDPM [32], and the same forward process formula means that our FEC-kv-reuse sampling adapts to the pre-trained models of DDPM.

The algorithm of three methods with inversion and sampling stages is provided in Algorithm 1, Algorithm 2, Algorithm 3.

## IV. EXPERIMENTS

**Experimental Setting**. We evaluate the quality of reconstruction and editing using the pre-trained Stable Diffusion model [33]. we create a total of 143 image-prompt pairs from Tedbench [34] and our own image set MixtureBench. MixtureBench cover images of animals and humans, some of which contain complex background images, low resolution and edges, which enhances the possibility of reconstruction distortion. All images are interpolated and scaled to RGB images of 512 x 512 size. The inference step for DDIM inversion and sampling is set to 50, because 50 steps is commonly used in business and is a preferable trade-off in terms of speed and quality.

In the reconstruction experiments, we compare our three methods with direct DDIM sampling(baseline), null-text inversion [31], Negative-prompt inversion [11], observe the reconstruction result under different guidance scale and different prompts, and analyse their sensitivity to batch size. In the editing experiments, we apply our 3 methods to state-of-the-art image editing methods, including Prompt2Prompt [7] and Masactrl [20]. In addition, We conduct an ablation study to see the effect of reusing only V. In conclusion, We will compare the reconstruction and editing quality, time consumption as well as GPU memory usage.

### A. Reconstruction

#### 1) Comparisons with Other Methods:
**Comparison Methods**. We compare our three methods with direct DDIM sampling (baseline), Null-text inversion [31] and Negative-prompt inversion [11]. Baseline is directly sampling from the final latent of DDIM inversion stage. Null-text inversion finetunes an unconditional text embedding for each step of the sampling, using the inversion trajectory as pivot. Negative-prompt inversion directly sets the unconditional text embedding to conditional text embedding in sampling stage, which is actually equivalent to setting the guidance scale to 1.0.

**Metrics**. We select objective metrics such as latent loss, PSNR and SSIM to measure the reconstruction performance. Latent loss refers to the mean squared error (MSE) between the initial latent(the output tensor of image through VAE encoder) of the inversion stage and the final latent of the sampling stage. A smaller value of latent loss indicates better reconstruction performance. Peak Signal to Noise Ratio(PSNR) measures the ratio between the maximum signal and the background

TABLE I: Average Reconstruction Metrics

| Method / Metric | direct(baseline) | Null-text(tuning) | Neg-prompt | FEC-kv-reuse | FEC-ref | FEC-noise |
|---|---|---|---|---|---|---|
| **Latent loss** ↓ | 0.4411 | 0.0018 | 0.0568 | 0.0033 | 6e-6 | **7e-16** |
| **PSNR** ↑ | 16.567 | 26.093 | 21.937 | 25.924 | **26.305** | **26.305** |
| **SSIM** ↑ | 0.527 | 0.737 | 0.661 | 0.737 | **0.741** | **0.741** |

Note: (1)PSNR and SSIM between source image and the output image of VAE encoding and then decoding are 26.305, 0.741. (2) Requirement: latent loss < 0.01; PSNR, SSIM approach to the value mentioned in (1)

noise. Larger PSNR indicates higher reconstruction performance. SSIM means the structural similarity between image. Higher SSIM value indicates higher similarity.

In general, it can be assumed that the difference between the reconstructed image and source image is hardly noticeable when the latent loss is less than 0.008. Stable diffusion utilizes VAE for conversion between image and latent space. VAE encoding and decoding may reduce the values of PSNR and SSIM, particularly for images with dense border lines since these may appear blurred by VAE. In consequence, we consider ideal score of PSNR and SSIM for reconstruction approach to that between original image and the output image of VAE decoding after encoding. It is noted in table I.

**Results**. Result of quantitative index is presented in Table I, and example images are in Figure 1. We calculate the average latent loss, PSNR, and SSIM scores for 143 image-prompt pairs across three different guidance scales(1.0, 5.0, 7.5). The guidance scale during the inversion stage and the sampling guidance scale are set to equivalent values.

We can observe that direct DDIM sampling results in significant reconstruction distortion. Its reconstructed images are often quite different from the source images, and its three metric scores fall far below the required levels. Its tent loss exceeds 0.45, and the PSNR falls below 17 and the SSIM only 0.5. Therefore, using the baseline method to edit real images might result in significant distortion, including identity, scene and textual distortion.

Our methods, FEC-ref and FEC-noise outperform the remaining methods in terms of these three metrics and can achieve lossless reconstruction satisfactorily. Their latent loss is below 10e-5,while the PSNR and SSIM score reach to the optimal level (level of the VAE). Null-text inversion achieves third place according to the metrics, however, it necessitates time-consuming fine-tuning (80-150 seconds). Our FEC-kv-reuse method also fulfills the requirements of reconstruction. It significantly outperforms Negative-prompt inversion, but is slightly inferior to the finetune-based null-text inversion method in terms of latent loss and PSNR. In conclusion, all three of our finetune-free methods are capable of achieving good reconstruction.

*2) Reasons of Our Reconstruction Success:*

Sec. III-B describes the accumulative error that causes reconstruction failure. All three of our methods substantially reduce the accumulative error by incorporating information from the inversion trajectory.

FEC-kv-reuse substitutes the key-value pairs of the self-attention during the sampling stage with the those in the corresponding step at the inversion stage, thus it reduces the error of predicted noise between the two stages. Figure 4 displays the trend of corresponding latent loss between reconstruction and inversion stages.The x-coordinate represents timesteps(not the number) used in reconstruction sampling stage, while the y-coordinate denotes corresponding latent loss between reconstruction and inversion stages. For direct DDIM sampling, latent loss grows exponentially with process. For FEC-kv-reuse, the loss increases at a very slow rate, always below 0.007.

FEC-ref and FEC-noise directly utilize each latent of the inversion phase. FEC-ref use latents of inversion stage as UNet input, FEC-ref can make each latent loss less than10e-5. Further, FEC-noise uses latents of inversion stage as the goal of the next step. Therefore,The only loss of FEC-noise method is the computer accuracy error, always below $10^{-15}$.

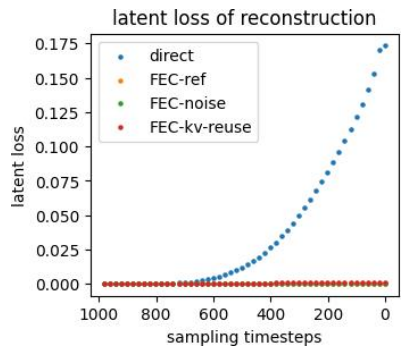

Fig. 4: Latent loss between reconstruction and inversion. We substantially reduce the accumulative error. 1)Direct DDIM sampling: grows exponentially with sampling process. 2)FEC-kv-reuse: always below 0.007; increases at a very slow rate. 3)FEC-ref, FEC-noise: always below $10^{-5}$, $10^{-15}$, respectively.

*3) Sensitivity to Guidance Scale:*

In this section, we discuss the effect of the guidance scale, mentioned in Eq. 4, on reconstruction. We focus on the following two conditions.

(a) Set same value to the guidance scale of inversion and sampling.

Results are displayed in Table II. It is apparent from the tables that the reconstruction outcome of direct DDIM sampling is frequently unacceptable (latent loss > 0.01 on average), particularly with a larger guidance scale (latent loss > 0.9). Regarding FEC-kv-reuse, the latent loss, PSNR, and SSIM show a slight decline as the guidance scale increases, but remain within an acceptable range (latent loss < 0.01, PSNR > 25). We note that reconstructed images with a lower quality outcome of FEC-kv-reuse can still preserve the structure and texture of the original image. In addition, both FEC-ref and FEC-noise maintain satisfactory reconstruction at any guidance scale.

Why direct DDIM sampling and FEC-kv-reuse are affected by guidance scale? According to Equation 4, the guidance scale controls the amplitude of the conditional and unconditional text embedding (corresponding to prompt and Negative-prompt, respectively). The greater the size of the guidance scale, the more powerful the prompt and neg prompt messages become for the editor, making it more difficult to sustain the original state of the reconstruction outcomes.

Why FEC-ref and FEC-noise are unaffected by guidance scale? Both FEC-ref and FEC-noise reference the inversion trajectory at each step, thus they can correct errors at each reconstruction step.

(b) Set different value to the guidance scale of inversion and sampling The higher the guidance scale, the more intensive the guidance message (prompt and neg prompt) to the editor, and therefore, sampling with a higher guidance is assumed to produce better editing results, such as 7.5 [31]. We fix the sampling guidance scale to 7.5 and conduct reconstruction experiments under different inversion guidance scale. Results are shown in Table III. Reconstruction of direct method fails miserably under any guidance, while FEC-kv-reuse requires an inversion guidance scale of 7.5 to meet our standard. In addition, both FEC-ref and FEC-noise maintain satisfactory reconstruction regardless of guidance, as they reference inversion trajectory at each step.

*4) Sensitivity to prompt type:*

We categorize our data according to prompt type: empty and non-empty. Table IV presents the outcomes. Reconstruction of baseline method performs worse in the non-empty case, revealing more significant loss of information. In contrast, the metrics of our three methods show similar values across different prompt types. This indicates that our approaches are almost unaffected by whether the prompt has meaning.

We provide an explanation. Note that conditional and unconditional text embedding are the same when both prompt and neg prompt are set to "" (null). According to Equation 4, at this point, no matter what guidance scale is set, it is equivalent to the case where guidance scale is 1. As a result, Stable Diffusion will be less affected by the prompt.

*5) Sensitivity to batch size:*

In Section III-B, we illustrate the problem of batch size and accumulative errors. All three of our methods solve this problem. Outcomes are displayed in Figure 3. FEC-kv-reuse inversion utilize identical batch size in inversion and editing, thereby avoiding problem of batch size. FEC-ref utilizes the latents from inversion directly at each sampling step, thereby significantly decreasing the accumulative error resulting from the batch size problem. FEC-noise takes a further step and directly makes the next step the same as the inversion, delivering the correct unconditional noise during editing stage. This avoids the problem of batch size and accumulative error as well.

## B. Editing: Consistency Maintenance

In this section, we apply our three methods to state-of-the-art image editing methods, including Prompt-to-Prompt [7] and Masactrl [20]. We compare the quality of editing and demonstrate that our methods maintain the consistency of the unedited parts in both rigid and non-rigid editing.

TABLE II: Sensitivity to guidance scale $\omega$

| Method/Guidance | Latent loss | | |
|---|---|---|---|
| | 1.0 | 5.0 | 7.5 |
| FEC-kv-reuse | 0.0010 | 0.0025 | 0.0083 |
| FEC-ref | 6e-6 | 6e-6 | 6e-6 |
| FEC-noise | 7e-16 | 7e-16 | 7e-16 |

| | PSNR ↑ | | | SSIM ↑ | | |
|---|---|---|---|---|---|---|
| Method/$\omega$ | 1 | 5 | 7.5 | 1 | 5 | 7.5 |
| direct | 21.937 | 12.687 | 11.081 | 0.661 | 0.425 | 0.395 |
| FEC-kv-reuse | 26.314 | 25.983 | 25.185 | 0.742 | 0.736 | 0.731 |
| FEC-noise | 26.325 | 26.287 | 26.287 | 0.742 | 0.739 | 0.739 |
| FEC-ref | 26.326 | 26.288 | 26.288 | 0.742 | 0.739 | 0.739 |

Note: Inversion guidance = Sampling guidance. FEC-kv-reuse remains within an acceptable range (latent loss < 0.01, PSNR > 25). FEC-ref and FEC-noise are unaffected by guidance scale.

TABLE III: Sensitivity to inversion guidance scale: latent loss

| Method \ Inv guidance | 1 | 3 | 5 | 7.5 |
|---|---|---|---|---|
| direct | 0.2357 | 0.3805 | 0.5516 | 0.9270 |
| FEC-kv-reuse | 0.0676 | 0.0886 | 0.0524 | 0.0084 |
| FEC-ref | 9e-6 | 8e-6 | 7e-6 | 6e-6 |
| FEC-noise | 7e-16 | 7e-16 | 7e-16 | 7e-16 |

Note: Fix sampling guidance = 7.5. Direct method fails miserably, while FEC-kv-reuse requires an inversion $\omega = 7.5$. FEC-ref and FEC-noise always maintain satisfactory reconstruction.

TABLE IV: Sensitivity to prompt type

| | empty prompt | | | non-empty prompt | | |
|---|---|---|---|---|---|---|
| Method | loss↓ | PSNR↑ | SSIM↑ | loss↓ | PSNR↑ | SSIM↑ |
| direct | 0.0737 | 20.93 | 0.643 | 0.5322 | 15.48 | 0.498 |
| FEC-kv-reuse | 0.0010 | 26.36 | 0.745 | 0.0039 | 25.81 | 0.735 |
| FEC-ref | 7e-6 | 26.37 | 0.756 | 6e-6 | 26.28 | 0.739 |
| FEC-noise | 7e-16 | 26.37 | 0.746 | 7e-16 | 28.28 | 0.739 |

Note: Our methods are almost unaffected by prompt type

*1) Rigid Editing:*

Rigid editing usually refers to the editing result has the same overall structure with the source image, that is to say, the SSIM of their edge maps is high. This type of editing includes object replacement, object adding, scene replacement, style transfer and so on. We apply our methods to Prompt-to-prompt [7] and complete such editing tasks.

Outcomes are presented in Figure 5 and Figure 7. The baseline refers to using the final latent DDIM inversion as the initial latent of Prompt-to-prompt. As the unsuccessful reconstruction conveys inaccurate information to the editing route, there appears inconsistency of unedited parts with the source image in the editing outcome. For instance, we only want to modify the scene, but the baseline method also alter the person's identity due to unsuccessful reconstruction.

In contrast, FEC-ref and FEC-noise can avoid accumulative error and reconstruct successfully, passing correct information (it is cross attention maps and latents in prompt-to-prompt) to the editing route. Therefore, our methods can make the unedited parts consistent with the source image. As shown in figure 7, we can change the clothes of two dogs and the background of images, while preserving the dogs' identity (the dogs in editing outcome is the same as the those in source image). In addition, in many cases, if FEC-ref and FEC-noise use the same parameter value of prompt-to-prompt, the results they produce exhibit significant resemblance, since they both use

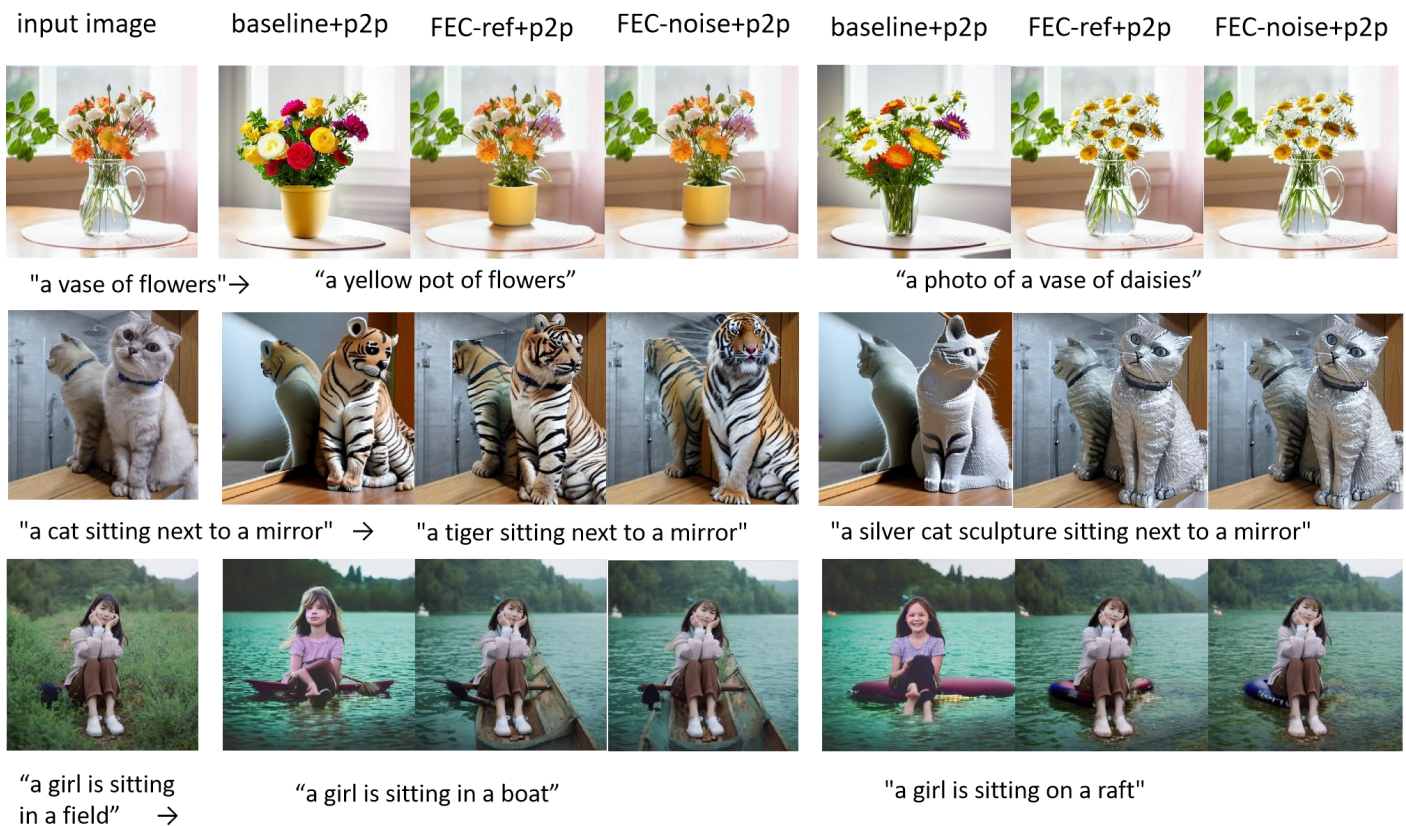

Fig. 5: Results of applying our methods to Prompt-to-Prompt (p2p) [7]. The three rows represent object replacement, object replacement, and scene replacement, respectively. Our methods assist to preserve the consistency of unedited parts, compared to the baseline.

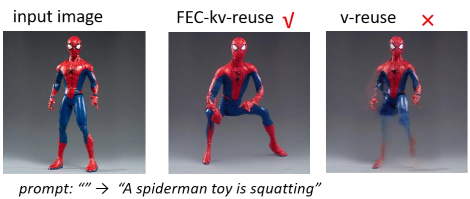

Fig. 6: Ablation study. Only reusing V cannot satisfy our editing requirements.

TABLE V: Ablation study: latent loss

| Method | Avg(ig=g) | ig=g=7.5 | ig=1,g=7.5 |
|---|---|---|---|
| FEC-kv-reuse | 0.0033 | 0.0084 | 0.0676 |
| FEC-v-reuse | 0.0043 | 0.0104 | 0.0343 |

Note: Ablation study, only reuse V vs FEC-kv-reuse, latent loss of reconstruction. *ig*=invertion guidance scale, *g*=sampling guidance scale. FEC-v-reuse can reduce major accumulative error of reconstruction, sometimes meeting our requirements (latent loss < 0.01).

TABLE VI: Time and Memory Usge

| Method-edit | Prompt-to-prompt [7] | | MasaCtrl [20] | |
| Method-rebuild | Time(s) | Mem(GB) | Time(s) | Mem(GB) |
|---|---|---|---|---|
| direct | 18.8 | 14.2 | 22.6 | 15.0 |
| Null-text inversion | 96~150 | 23.0 | 96~150 | 23.5 |
| FEC-kv-reuse | \ | \ | **15.0** | 15.0 |
| FEC-ref | 18.8 | 14.2 | 22.6 | 15.0 |
| FEC-noise | 18.8 | 14.2 | 22.6 | 15.0 |

Note: The time and peak GPU memory used to edit a real image. Our methods, without fine-tuning, produce comparable editing results to the Null-text inversion [31], but are **five times faster** and use only 60% of the GPU memory.

the reference the inversion trajectory.

In conclusion, we can realize consistent real image editing, including object replacement, object adding, scene replacement and so on.

*2) Non-Rigid Editing:*

Non-rigid editing involves modifying the structure of an object, such as editing posture(action) of animals and people.

We apply our methods to MasaCtrl [20] to perform action editing.

Outcomes are presented in Fig 8. It appears that MasaCtrl may lose individual identity and texture or produce a blurry image as a result of unsuccessful reconstruction. In contrast, all of our methods can reconstruct successfully, passing correct information ( it is self attention KV in MasaCtrl, containing texture and some structure) to the editing route. For instance, we can make the running dog in editing result quite resemble the dog in source image.

*C. Ablation Study: Only Reuse V*

Our FEC-kv-reuse, which eliminate the reconstruction process, is designed for situations like MasaCtrl where the editing use the KV from reconstruction.

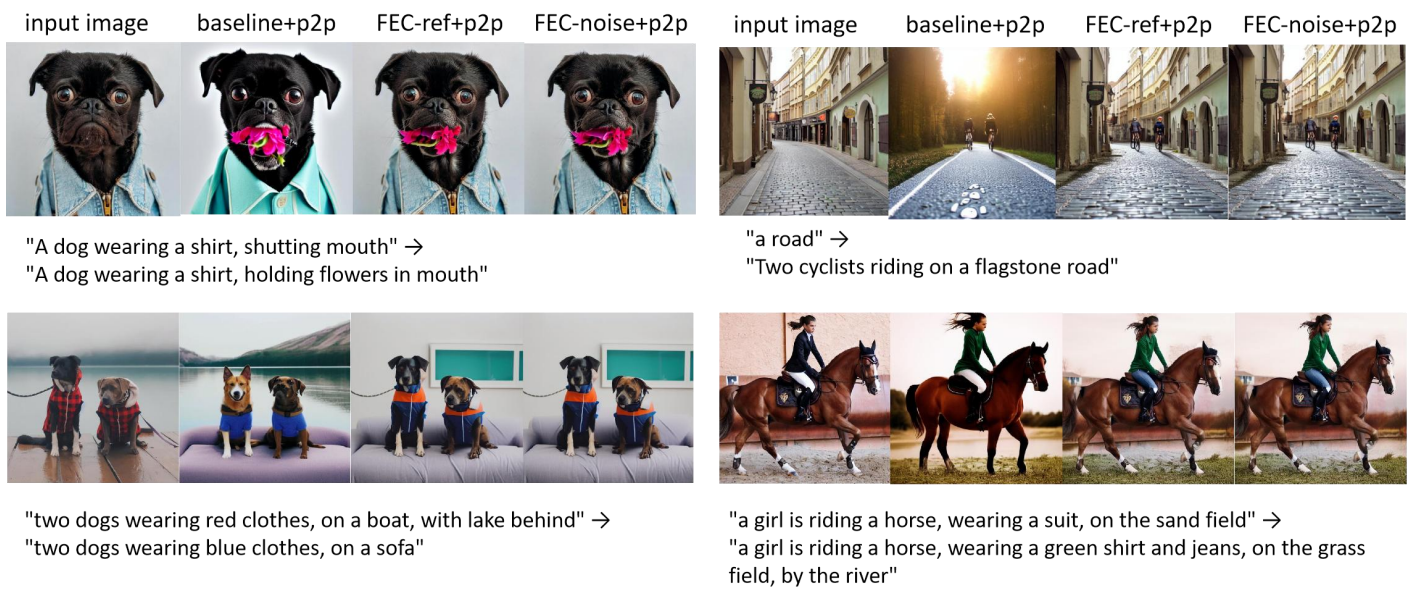

Fig. 7: Results of applying our methods to Prompt-to-Prompt (p2p) [7]. The two rows represent object adding, object and scene replacement, respectively. Our methods assist to preserve the consistency of unedited parts, compared to the baseline.

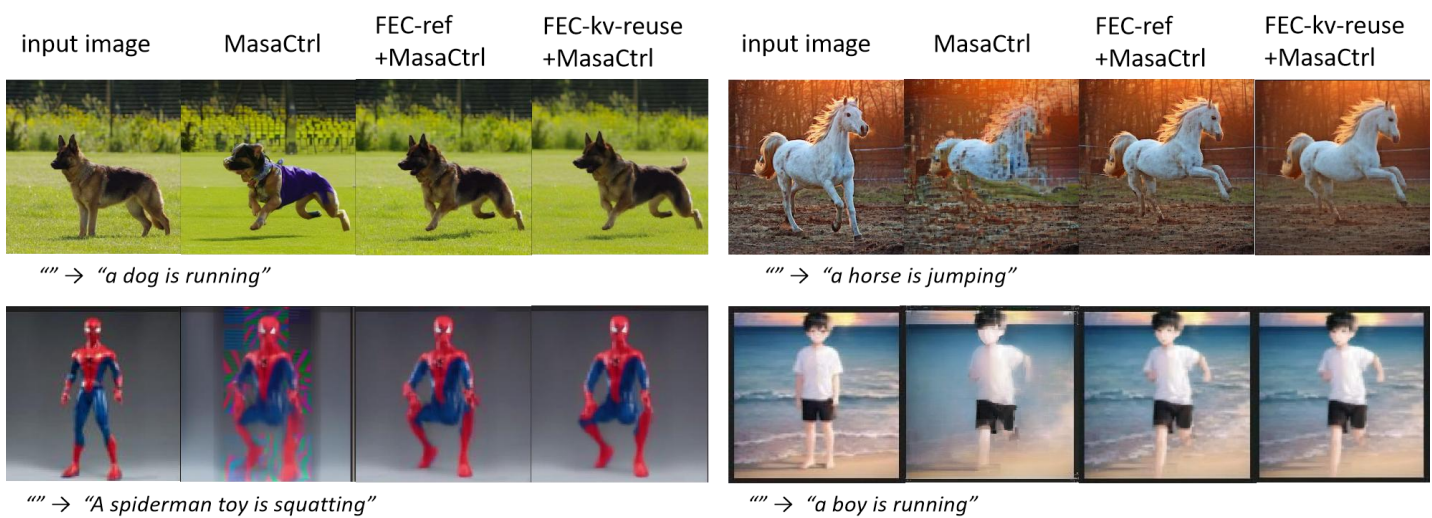

Fig. 8: Results of applying our methods to MasaCtrl [20], a non-rigid editing technique. The original MasaCtrl may loss individual identity and texture or produce blurry images, due to distorted reconstruction. Our methods assist to solve these problems and preserve the consistency of unedited parts.

In order to explore the effect of reusing V on the elimination of accumulative errors during reconstruction, we cancel reusing K and test the reconstruction metrics. As shown in the TableV, we observe that the performance of FEC-v-reuse and FEC-kv-reuse are similar. Reusing only V can greatly reduce the accumulative error in the reconstruction process, meeting our requirements (loss <= 0.01) sometimes. In addition, FEC-v-reuse can save a portion of the GPU memory compared to FEC-kv-reuse.

However, only reusing V cannot accomplish the editing tasks, as shown in Fig6. We generally require K and V to come from the same place when we use attention mechanism. If we only reuse V in editing, k and v will not correspond, thus confuse the latent space.

### D. Time and Memory Usage

We record the time and peak GPU memory usage required to complete an edit with our methods, as presented in Table VI. Our experiments are conducted on a NVIDIA RTX A6000.

Our FEC-kv-reuse is the fastest. In terms of memory usage memory usage, our three methods utilize approximately the same amount as the direct method. Null-text inversion [31] is the most time-consuming and memory-consuming as it requires training unconditional text embedding tensor. The training process demands around 19GB of GPU memory and takes approximately 90-150 seconds. Compared to the tuning-based method, our three methods are five times faster and use approximately 60% of the GPU memory, while producing comparable editing results.

We explain why FEC-kv-reuse is the fastest. In the original Prompt-to-Prompt [7] and MasaCtrl [20], the editing stage involve two parallel routes: reconstruction and editing. By eliminating the reconstruction route, kv-reuse can halve the time required for the editing stage. Additionally, FEC-kv-reuse also decreases GPU usage during editing; however, it requires storing self-attention KV at the inversion stage, increasing GPU memory consumption. These opposing effects balance each other out, so FEC-kv-reuse utilizes similar amount of memory as the direct method, FEC-ref and FEC-noise.

## V. Limitations and Conclusion

In this paper, we propose FEC, which consists of three sampling methods that guarantee perfect reconstruction performance under different settings. We have greatly improved reconstruction and editing performance in image editing tasks. Besides, compared to training- or fine-tuning-based methods, our approach guarantees reconstruction performance extremely quickly and greatly reduces unnecessary cost of time as well as the use of computer memory.

Of course, our method has drawbacks, due to the lack of training and fine-tuning, our method needs to rely on downstream editing methods to complete the editing task, and thus the main contribution of our method is currently on improving the reconstruction performance. Our future work will focus more on how to design more complex and diverse editing techniques to propose more novel editing tasks.